\documentclass[11pt]{article}

\usepackage[final]{acl}

\usepackage{times}
\usepackage{latexsym}

\usepackage[T1]{fontenc}

\usepackage[utf8]{inputenc}

\usepackage{microtype}

\usepackage{inconsolata}

\usepackage{graphicx}

\usepackage{booktabs}
\usepackage{hyperref}
\usepackage{tabularx}

\usepackage{fontawesome5}

\usepackage{booktabs, colortbl} 
\usepackage{multirow,makecell} 
\usepackage{multicol}

\usepackage{amsmath}
\usepackage{amssymb}
\usepackage{caption}
\usepackage{subcaption}
\usepackage{titlesec}
\usepackage{pgfplots}
\usepackage[normalem]{ulem} 
\usepackage{enumitem}
\usepackage{graphicx} 
\usepackage{hyperref}
\usepackage{svg}

\usepackage{enumitem}
\usepackage[splitrule]{footmisc}
\setlist[itemize]{noitemsep, nolistsep}
\usepackage{color, colortbl}

\title{Conditions for Catastrophic Forgetting in Multilingual Translation}

\author{Danni Liu \quad\quad Jan Niehues \\
        Karlsruhe Institute of Technology, Germany \\
        \texttt{\{danni.liu, jan.niehues\}@kit.edu}} 

\begin{document}
\maketitle
\begin{abstract}
Fine-tuning multilingual foundation models on specific languages often induces catastrophic forgetting, degrading performance on languages unseen in fine-tuning. 
While this phenomenon is widely-documented, 
the literature presents fragmented results about when forgetting occurs.
To address this ambiguity, 
we conduct a systematic empirical study using machine translation as a testbed to identify the conditions that trigger catastrophic forgetting in multilingual fine-tuning. 
Through controlled experiments across different model architectures, data scales, and fine-tuning approaches, 
we reveal that the relative scale between model and data size is a primary determinant of forgetting. 
Moreover, we demonstrate that a model's instruction-following ability is more critical for retaining multilingual knowledge than its architecture. 
Contrary to assumptions, parameter-efficient fine-tuning offers no clear advantage over full fine-tuning in mitigating forgetting. 
Lastly, we show that cross-lingual alignment can mitigate forgetting while also facilitating positive transfer to unseen target languages.
\end{abstract}

\section{Introduction}
Foundation models pretrained on vast amounts of multilingual data have become the standard backbone for modern natural language processing systems.
To achieve optimal performance, however, these models typically require fine-tuning on downstream tasks. 
This specialization introduces a critical trade-off: 
while performance on the target task improves, the model may suffer from \textit{catastrophic forgetting} \cite{MCCLOSKEY1989109}, 
a substantial degradation of capabilities on tasks or languages not present in the fine-tuning data.

\begin{figure}[t]
    \centering
    \includegraphics[width=0.45\textwidth,clip,trim={0 2.9cm 0 3.4cm}]{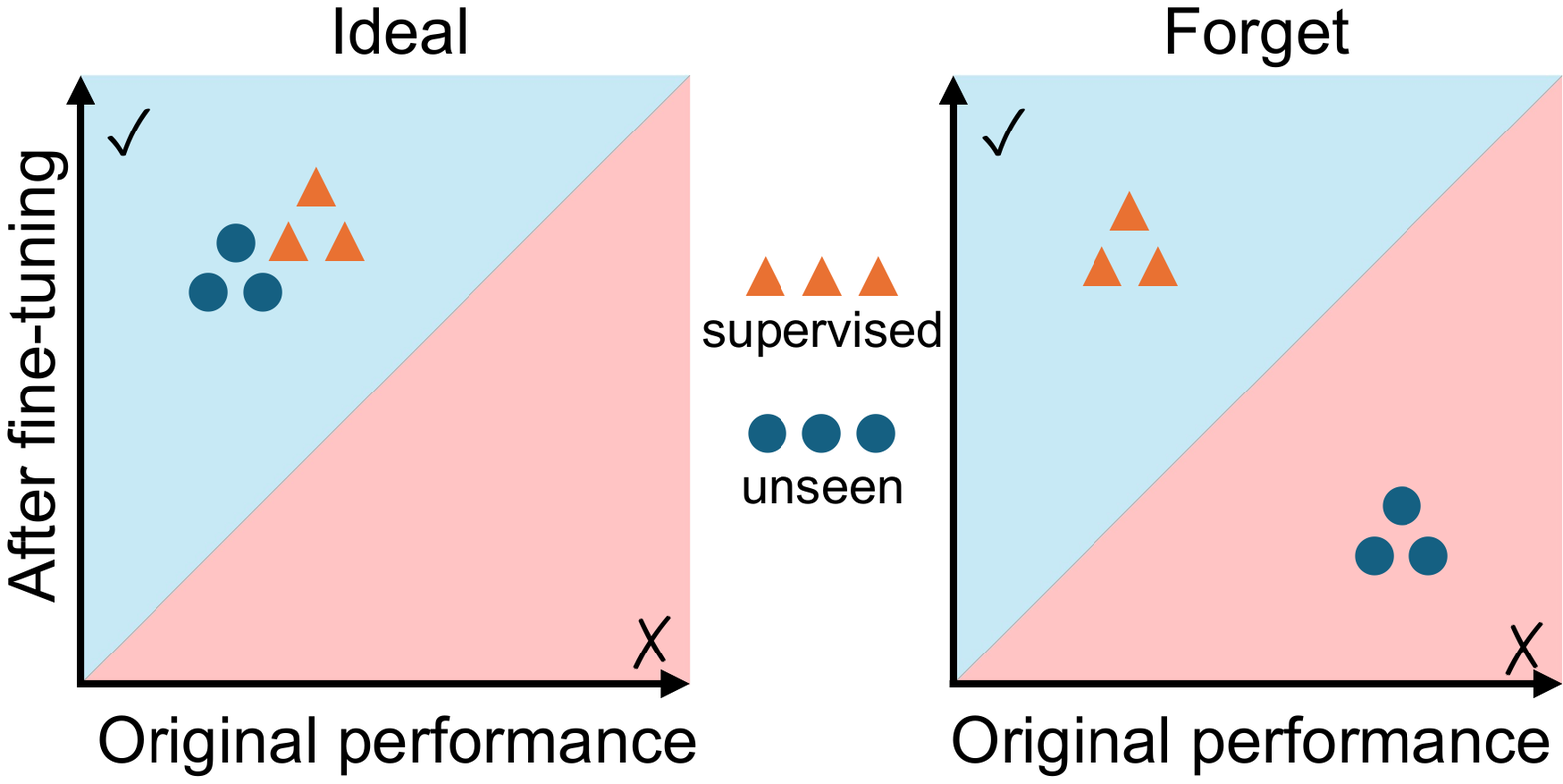}
    \caption{Selectively fine-tuning on some languages or translation directions may lead to positive transfer (left) or catastrophic forgetting (right).}
    \label{fig:when_forgetting_happens_overview}
\end{figure}

A common use case is to fine-tuning multilingual models to focus on specific languages or language pairs. 
Ideally, this process would not harm, and might even improve, performance on unseen languages through positive transfer, as illustrated on the left of \autoref{fig:when_forgetting_happens_overview}.
However, empirical evidence often shows the opposite. 
Models frequently lose proficiency in languages they were not fine-tuned on \cite{vu-etal-2022-overcoming,sun-etal-2023-efficiently,winata-etal-2023-overcoming}, 
as shown on right of \autoref{fig:when_forgetting_happens_overview}.

Machine translation (MT) serves as a compelling testbed for studying multilingual catastrophic forgetting. 
First, a model supporting $n$ languages encompasses $n(n-1)$ directed translation pairs, 
offering a large and structured space to analyze forgetting patterns. 
Second, languages can be ``unseen'' in different roles. 
For example, 
a language may be present only as a source language, 
only as a target language, 
or in specific source-target pairs that were never explicitly trained.
This enables fine-grained analysis of how different types of exposure during fine-tuning affect retention.
Moreover, forgetting can occur asymmetrically, 
where a model may retain the ability to translate from language A to B while losing the reverse direction.

Despite its practical importance, 
the literature presents a fragmented and sometimes contradictory picture of when catastrophic forgetting occurs in MT.
On one hand, 
studies on traditional NMT models trained from scratch \cite{berard-2021-continual} and some large pretrained models \cite{vu-etal-2022-domains,liu-niehues-2022-learning,liu-etal-2023-continual,lai2023extendingmultilingualmachinetranslation} report severe forgetting after standard fine-tuning, 
where the ability to translate unseen directions is almost entirely lost. 
These findings suggest that catastrophic forgetting is an inevitable consequence of selective specialization.
On the other hand, recent works on large language models (LLMs) provided mixed evidence.
\citet{richburg2024how} demonstrated that fine-tuning Llama 2 \cite{DBLP:journals/corr/abs-2307-09288} and Tower \cite{alves2024tower} models on specific language pairs could improve performance on unseen pairs, indicating positive transfer.
Conversely, \citet{zan2024buildingaccuratetranslationtailoredllms} found that their fine-tuned Llama 2 models performed very poorly on unseen directions, 
again indicating issues with forgetting.

These conflicting results raise fundamental questions about the factors leading to catastrophic forgetting. 
Has the emergence of large language models altered the dynamics of catastrophic forgetting? 
To what extent do model architecture (encoder-decoder versus decoder-only), 
scale, 
or fine-tuning methodology determine whether a model forgets or generalizes?
How do factors like the volume of fine-tuning data, the use of parameter-efficient fine-tuning (PEFT), or instruction-following capabilities influence the retention of multilingual abilities?
To resolve these ambiguities, 
we conduct a systematic study to identify the conditions that trigger catastrophic forgetting in multilingual MT.
We systematically control for key variables, including
model architecture and size, 
fine-tuning data composition and scale, 
full-parameter vs. parameter-efficient fine-tuning, 
and instruction-following versus standard fine-tuning approaches.
With a series of controlled experiments, we demonstrate that:
\begin{itemize}[nolistsep,leftmargin=*]
    \item  
    The relative scale between pre-trained model parameters  and fine-tuning data volume is a critical factor in catastrophic forgetting, with smaller models fine-tuned on larger datasets being most vulnerable  (\S\ref{subsec:chap6_model_data_relative_size}).
    \item Whether a model supports instruction-following, rather than its underlying architecture (encoder-decoder versus decoder-only), is a primary factor impacting catastrophic forgetting (\S\ref{subsec:chap6_architecture_and_lang_control}).
    \item Contrary to common assumptions, parameter-efficient fine-tuning with LoRA \cite{lora} provides no significant advantage over full fine-tuning in preventing catastrophic forgetting under our experimental conditions (\S\ref{subsec:lora_vs_full_FT}).
    \item Besides mitigating forgetting, cross-lingual alignment methods may facilitate positive transfer,
    with improvements observed on translation directions with unseen target languages
    (\S\ref{sec:chap6_cross_lingual_align}).
\end{itemize}

\section{Related Work}
\paragraph{Catastrophic Language Forgetting in MT}
Catastrophic forgetting in machine translation has been extensively studied.
\citet{dakwale-monz-2017-fine,thompson-etal-2018-freezing,thompson-etal-2019-overcoming} established that domain-specific fine-tuning degrades performance on previously learned domains with specific subject areas or text styles.
Many subsequent works have investigated the underlying mechanisms and mitigation strategies for domain forgetting in MT,
e.g., \citet{gu-feng-2020-investigating, saunders-deneefe-2024-domain, eschbach-dymanus-etal-2024-exploring,wu-etal-2024-f,hu-etal-2024-large-language}.
Compared to domain forgetting,
the multilingual dimension of forgetting has received less attention. 
\citet{berard-2021-continual} demonstrated severe language forgetting in conventional encoder-decoder-based MT models during standard fine-tuning on selected languages,
while \citet{vu-etal-2022-domains} showed that domain-specific fine-tuning compounds forgetting across both domains and languages.
\citet{liu-niehues-2022-learning,liu-etal-2023-continual} confirmed that standard fine-tuning consistently triggers catastrophic forgetting of unseen language pairs, even in pre-trained models with large language coverage, such as M2M-124 \cite{10.1162/tacl_a_00474} and mBART-50 \cite{tang-etal-2021-multilingual}.
Besides language and domain forgetting, 
models also lose in-context learning abilities after fine-tuning \citep{alves-etal-2023-steering}.

\paragraph{Model Factors Influencing Forgetting}
The scale of both the model and its pretraining data has been identified as a key factor in mitigating catastrophic forgetting \cite{ramasesh2022effect}. 
Our study extends this analysis by examining the relative scale between the model and the fine-tuning data. 
The choice of fine-tuning methodology is another contested factor. \citet{DBLP:journals/corr/abs-2401-05605} suggests that LoRA does not resolve catastrophic forgetting, while \citet{biderman2024lora} suggest that LoRA ``learns less and forgets less''. 
The finding by \citet{zhang2024when} that the optimal fine-tuning method is highly task-dependent warrants a specific investigation for the task of multilingual MT.

\section{Controlled Setting to Study Catastrophic Forgetting in MT}
Our controlled experiments are structured along two dimensions,
namely the choice of base model and the characteristics of the training dataset.

\subsection{Base Models}
An overview of all base models and their configurations is provided in~\autoref{tab:chap6_models}. 
\begin{table}[t!]
    \small
    \centering
    \setlength\tabcolsep{6pt}
    \begin{tabular}{l c c c c c c ccc}
    \toprule
    \textbf{Model} 
    & \textbf{Size}
    & \textbf{Model Type} 
    \\
    \midrule
    M2M-124-0.2B &
    175M &
    \multirow{2}{*}{Translation-specific}
    \\
    M2M-124-0.6B &
    615M &
    \\
    Qwen2.5-0.5B-Instruct &
    494M &
    \multirow{3}{*}{Instruction-following}
    \\
    Qwen2.5-7B-Instruct &
    7B
    \\
    Llama-3-8B-Instruct &
    8B
    \\
    \bottomrule
    \end{tabular}
    \caption{Base models and their configurations.\label{tab:chap6_models}}
    \vspace{-10pt}
\end{table}

\paragraph{Translation-Specific Models}
We choose M2M-124\footnote{We choose M2M-124 over NLLB-200 models of similar sizes \cite{nllb200} 
as the former showed stronger performance in our preliminary experiments.} 
\cite{10.1162/tacl_a_00474} with two sizes:
\begin{itemize}[nolistsep,leftmargin=*]
\item M2M-124-0.2B: smallest-scale baseline 
\item M2M-124-0.6B: larger-scale comparison to isolate model size effects
\end{itemize}

\paragraph{Instruction-Following Models}
We evaluate and fine-tune models from two prominent families, 
Qwen 2.5 \cite{qwen2025qwen25technicalreport} and Llama~3 \cite{grattafiori2024llama3herdmodels}:
\begin{itemize}[nolistsep,leftmargin=*]
    \item Qwen2.5-0.5B-Instruct : similar to M2M-124-0.6B in size for comparison between translation-specific and instruction-following models\footnote{We note that this is not fully controlled setup contrasting M2M-124-0.6B due to different pre-training data.}
    \item Qwen2.5-7B-Instruct: larger-scale instruction-following baseline
    \item Qwen2.5-7B-Instruct (LoRA): identical to full fine-tuning but using LoRA as a PEFT approach
    \item Llama-3-8B-Instruct (LoRA): similar scale to above but from another family
\end{itemize}









\subsection{Data}\label{subsec:data}

\begin{table}[t]
\centering
\small
\setlength\tabcolsep{4pt}
\begin{tabular}{l|p{6cm}}
\toprule
\textbf{Dataset} & \textbf{Details} \\
\midrule
\multirow{11}{*}{\textsc{small}} 
& \textbf{Training Data:} ALMA (117K sentence pairs) \\
& \textbf{Test}  (supervised): WMT23 \cite{kocmi-etal-2023-findings} \\
& \textbf{Test}  (unseen pair): WMT24 \cite{kocmi-etal-2024-findings} \\
& \textbf{Test}  (unseen source): WMT23 \\
& \textbf{Test}  (unseen target): WMT23 \\
& \textbf{Training directions:} \{cs, de, is, ru, zh\}$\leftrightarrow$en \\
& \textbf{Testing directions:} \\
& \quad - Unseen pair (20): \{cs, de, is, ru, zh\}$\leftrightarrow$\{cs, de, is, ru, zh\} \\
& \quad - Unseen source (3): \{he, ja, uk\} $\rightarrow$ en \\
& \quad - Unseen target (3): en $\rightarrow$ \{he, ja, uk\}  \\
\midrule
\multirow{10}{*}{\textsc{large}} 
& \textbf{Training Data:} WMT21 large-scale multilingual track (54M sentence pairs) \\
& \textbf{Test}  (unseen pair): FLoRes \cite{10.1162/tacl_a_00474} \\
& \textbf{Test}   (unseen source): FLoRes \\
& \textbf{Test}  (unseen target): FLoRes \\
& \textbf{Training directions:} \{jv, ms, tl\}$\leftrightarrow$en \\
& \textbf{Unseen testing directions:} \\
& \quad - Unseen pair (6): \{jv, ms, tl\}$\leftrightarrow$\{jv, ms, tl\} \\
& \quad - Unseen source (4): id $\rightarrow$ \{en, jv, ms, tl\} \\
& \quad - Unseen target (4): \{en, jv, ms, tl\} $\rightarrow$ id \\
\bottomrule
\end{tabular}
\caption{Dataset overview for training and testing configurations for both small and large-scale experiments.}
\label{tab:dataset_overview}
\vspace{-10pt}
\end{table}

\paragraph{Dataset Overview}
As shown in \autoref{tab:dataset_overview}, we experiment on datasets of different scales:
\begin{itemize}[nolistsep,leftmargin=*]
    \item \textsc{small}: training dataset in ALMA \cite{alma}, covering five languages paired with English: Czech (cs), German (de), Icelandic (is), Russian (ru), and Chinese (zh). 
    The unseen languages include Hebrew (he), Japanese (ja), and Ukranian (uk).
    \item \textsc{large}: from the WMT 21 Shared Task on Large-Scale Multilingual Machine Translation \cite{wenzek-etal-2021-findings},  focusing on three  related Austronesian languages paired with English: Javanese (jv), Malay (ms), and Tagalog (tl). The unseen language is Indonesian (id).
    \item subsampled \textsc{large}: sampled from the \textsc{large} dataset with 12K, 120K, and 1.2M sentences per language pair respectively.
\end{itemize}

\paragraph{Unseen Language Pairs} We evaluate catastrophic forgetting on three types of unseen language pairs. 
Our analysis focuses on pairs where at least one language was seen during fine-tuning, 
as pairs with two unseen languages consistently showed severe performance degradation in preliminary experiments. 
The three categories are:
\begin{itemize}[nolistsep,leftmargin=*]
\item \textbf{Unseen Pair}: Both the source and target languages are present in the fine-tuning data, but not in combination. This is the most challenging category as explained next.
\item \textbf{Unseen Source}: The source language has not been seen during fine-tuning, but the target language has.
\item \textbf{Unseen Target}: The target language has not been seen during fine-tuning, but the source language has.
\end{itemize}
Among the three evaluated categories, 
the ``unseen pair'' scenario presents a unique challenge.
While counterintuitive, 
this case is often more difficult than scenarios involving languages completely unseen during fine-tuning.
The primary reason for this difficulty lies in the English-centric nature of the fine-tuning dataset.
Because all training examples are paired with English, 
the model learns an implicit association that a specific source language uniquely predicts English as the target language, which represents a spurious correlation \cite{gu-etal-2019-improved}.\footnote{For instance, when translating from German-Czech after fine-tuning on English-Czech and German-English, 
the model has been implicitly trained to associate German inputs with English outputs.
Direct German-Czech translation requires the model to override this spurious correlation.}
In contrast, the other two conditions do not present this conflict to the same level.
For an unseen source language, 
the model has not formed any directional association during fine-tuning. 
Therefore, there is no learned association to be overridden.
The unseen target language scenario is also comparatively less difficult.
Specifically, as long as the source language has to translate into multiple different target languages during training or has not been seen in training, the model does not learn a one-to-one mapping to a single output.
This condition applies to four of the seven unseen target language scenarios (en $\rightarrow$\{he, ja, uk, id\}), 
where the source language was part of a multi-target translation setup.
This configuration discourages over-specialization toward a single output language, 
reducing the overall difficulty of translating into a unseen target language for this category.

\paragraph{Language Control Mechanisms}
Following the original models, we use different language specification methods.
For M2M-124, we follow their token-based control, prepending source and target sentences with their respective language tokens:
\begin{center}
\texttt{<source\_lang\_token> source sentence <target\_lang\_token> target sentence}
\end{center}
For instruction-following models, we use the system prompt ``Translate the given sentence from [source language] to [target language]'' followed by the source sentence. 
In ablations, we also test instructions in the target language\footnote{We translate English instructions with DeepL. For languages not supported by DeepL, we use Google Translate.}.

\subsection{Training and Inference}
For full fine-tuning, we update all model parameters.
For LoRA, we adopt a rank of 8 and $\alpha$ of 16,  applying adapters to all components within self-attention (Query, Key, Value, Output, Gate) and linear projections. 
This LoRA configuration was chosen after initial experiments applying LoRA to fewer components showed weaker supervised performance.
It also creates conditions more analogous to full fine-tuning than selective adapter application,  minimizing potential confounding factors related to parameter coverage.
More training and inference details are available in \autoref{sec:appendix_training_details}.

\begin{figure*}[t!]
    \centering
    \includegraphics[width=\linewidth,clip,trim={0 1.4cm 0 1cm}]{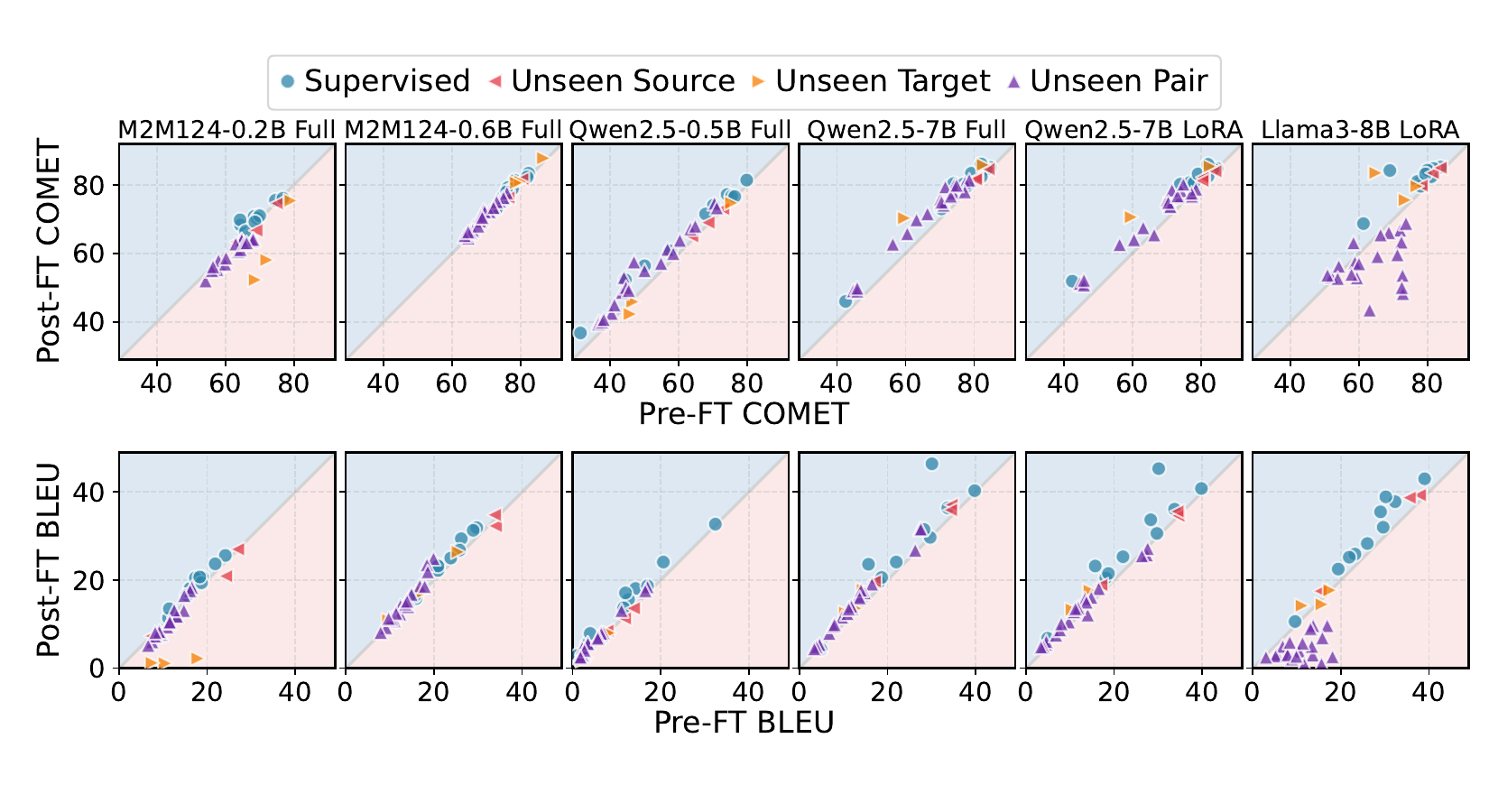}
    \caption{Gain-forgetting plots on the \textsc{small} dataset (117K sentence pairs). Catastrophic forgetting is minimal, except unseen language pairs on Llama (addressed later in \autoref{tab:chap6_inlang_instruct}). }
    \label{fig:gain_forget_plot_alma}
\end{figure*}

\begin{figure*}[t!]
    \centering
    \includegraphics[width=\linewidth,clip,trim={0 1.4cm 0 2.1cm}]{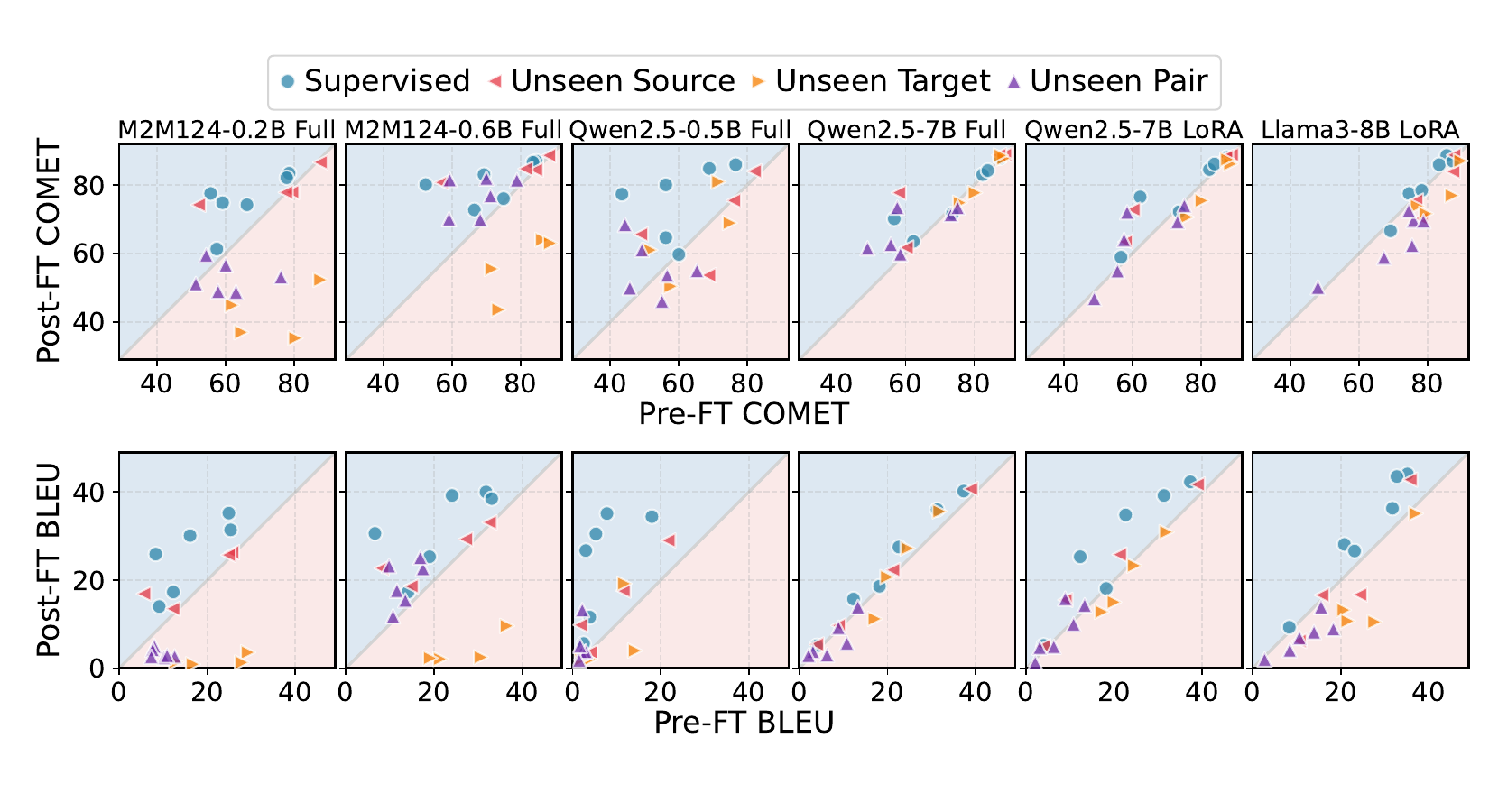}
    \caption{Gain-forgetting plots on the \textsc{large} dataset (54M sentence pairs). Catastrophic forgetting is more severe, especially with translation-specific models where they show performance collapse approaching 0 BLEU.}
    \label{fig:gain_forget_plot_wmt21}
\end{figure*}

\subsection{Metrics}
For evaluation, we primarily use COMET-22 \cite{rei-etal-2022-comet} as our main quality metric due to its strong correlation with human judgments \cite{freitag-etal-2022-results}.
However, COMET has known limitations when models generate unintended languages  \cite{zouhar-etal-2024-pitfalls}, which is particularly relevant for catastrophic forgetting.
Therefore,
we include BLEU\footnote{with default tokenizer ``13a'' in sacreBLEU \cite{post-2018-call}, and the dedicated tokenizers for Chinese and Japanese.} \cite{papineni-etal-2002-bleu} as a complementary string-matched metric.
When appropriate, we also report language accuracy using the language identification tool by \citet{lui-baldwin-2011-cross}.


\section{Gain-Forgetting Analyses}
We investigate the trade-off between performance gains on 
fine-tuned language pairs and potential catastrophic forgetting on those unseen during fine-tuning. 
To visualize this relationship, 
we create scatter plots (\autoref{fig:gain_forget_plot_alma} and \autoref{fig:gain_forget_plot_wmt21}) 
where each point represents a language pair's performance before ($x$-axis) and after ($y$-axis) fine-tuning. 
The diagonal line ($y=x$) is a reference boundary,
where points below indicate catastrophic forgetting, 
while those above indicate performance improvement.

\subsection{Model Scale and Fine-Tuning Data Size} \label{subsec:chap6_model_data_relative_size}
\paragraph{Impact of Model Size}
Larger model variants consistently exhibit greater resistance to catastrophic forgetting.
For M2M-124 models, 
the 0.6B parameter variant shows fewer language pairs in the forgetting zone compared to its 0.2B counterpart. 
Similarly for Qwen2.5, the 7B model demonstrates substantially less forgetting than the 0.5B model across all language pairs.
This confirms the finding from \citet{ramasesh2022effect} that the base model scale helps mitigate forgetting.

\paragraph{Impact of Fine-Tuning Data Volume}
We additionally observe that the amount of fine-tuning data plays a crucial role in forgetting.
By contrasting  \autoref{fig:gain_forget_plot_alma}  ($\sim$100K sentences FT data) and \autoref{fig:gain_forget_plot_wmt21} ($\sim$ 54M sentences FT data), 
it becomes clear that higher-data-volume fine-tuning leads to stronger forgetting across all model variants.
This observation extends the findings of \citet{ramasesh2022effect}, 
by demonstrating that catastrophic forgetting is impacted not only by base model scale, 
but also by the intensity of task-specific training.

\begin{figure}[t!]
    \centering
    \includegraphics[width=\linewidth,clip,trim={0 1.5cm 0 1cm}]{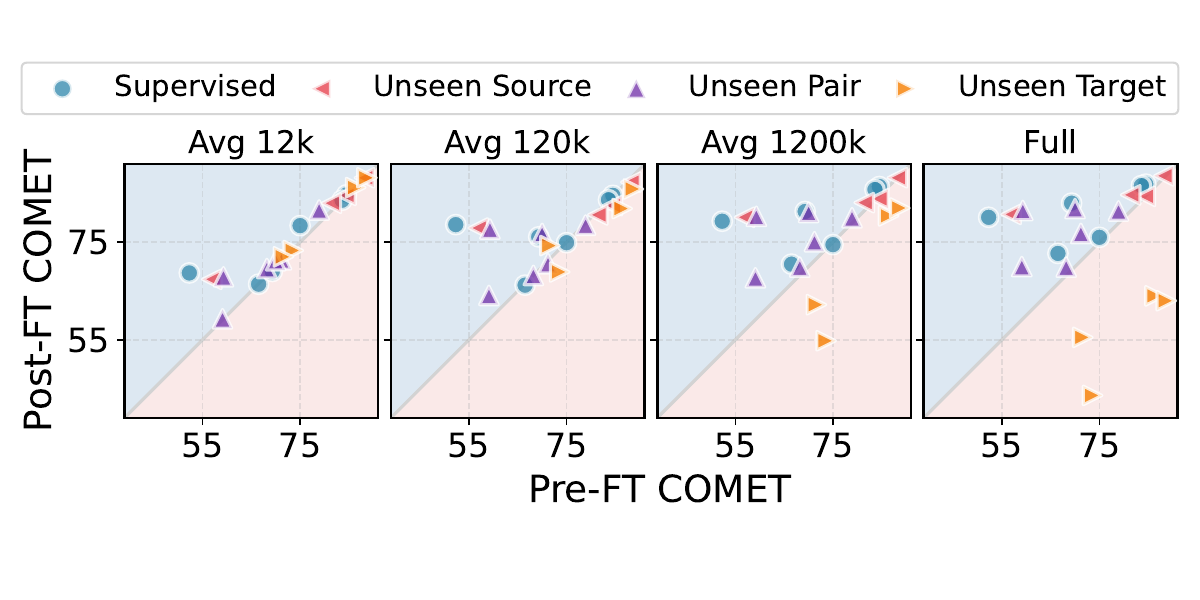}
    \caption{Controlled analysis of data volume effects by subsampled portions of the \textsc{large} dataset. 
    Forgetting becomes severe as fine-tuning data amount increases.}
    \label{fig:forgetting_wmt_subsample}
\end{figure}
\paragraph{Controlled Analysis of Data Volume Effects}
To isolate the impact of data volume from dataset-specific factors 
(e.g., the ALMA dataset has higher-quality data),
we conduct controlled experiments using subsampled portions of the WMT21 dataset. 
We systematically vary the amount of fine-tuning data while maintaining a consistent data source.
Starting with 12K sentences per language pair (matching ALMA), 
we increase the volume by an order of magnitude at each step: 12K $\rightarrow$ 120K $\rightarrow$ 1,200K sentences per language pair. 
\autoref{fig:forgetting_wmt_subsample} demonstrates the progressive increase in catastrophic forgetting as training volume grows.
At 12K sentences per language pair, the gain-forgetting pattern resembles ALMA results, 
with most language pairs clustered near the diagonal line and minimal performance changes. 
At 120K sentences, a shift toward forgetting emerges, particularly for target languages unseen in fine-tuning. 
At 1,200K sentences, severe catastrophic forgetting occurs, 
though not to the full extent observed in the complete dataset.
The progressive degradation suggests that the intensity of fine-tuning on specific language pairs impacts the forgetting patterns.




\subsection{Architecture and Language Control Mechanism} \label{subsec:chap6_architecture_and_lang_control}
To isolate the impact of model architecture and pretraining objectives on catastrophic forgetting, 
we compare two models of comparable scale but different designs: 
M2M-124-0.6B, a translation-specific encoder-decoder model,
and Qwen2.5-0.5B, a general-purpose decoder-only model pretrained for instruction following.
While these two models differ in their pre-training data, 
which preludes a fully controlled comparison,
we also replicate M2M-124's language control mechanism on Qwen2.5 to reduce potential confounding factors.

\paragraph{Forgetting Patterns Across Architectures}
In \autoref{fig:gain_forget_plot_wmt21} on the \textsc{large} dataset, where forgetting effects are strongest due to large-scale fine-tuning data, 
both models exhibit catastrophic forgetting with multiple language pairs falling below the diagonal. 
However, they differ in their forgetting patterns: 
M2M-124-0.6B exhibits severe performance degradation on unseen target languages, 
while Qwen2.5-0.5B shows modest forgetting.
We hypothesize that this is related to the target language control mechanisms by the models. 
As discussed in \S\ref{subsec:data}, M2M-124 relies on language-specific tokens prepended to both source and target sentences, with the target-side token determining the output language.
In contrast, with Qwen, we use natural language instructions to specify the target language, leveraging its existing instruction-following capabilities.
This instruction-based mechanism may support more generalizable language control and help mitigate catastrophic forgetting on unseen target languages.
We examine this hypothesis next.

\begin{table}[t!]
    \small
    \centering
    \setlength\tabcolsep{2pt}
    \begin{tabular}{l c c c c c c ccc}
    \toprule
    \multirow{2}{*}{\textbf{Model}}
    & \multicolumn{2}{c}{\textbf{Average}}
    &
    & \multicolumn{2}{c}{\textbf{en$\rightarrow$id}}
    \\
    \cmidrule{2-3}
    \cmidrule{5-6}
    & COMET
    & BLEU 
    &
    & COMET
    & BLEU
    \\
    \midrule
    Qwen2.5-0.5B-Instruct &
    64.0 &
    \textbf{8.0} &
    &
    71.6 &
    11.8
    \\
    + Instruction-based FT &
    \textbf{65.3} &
    6.8 &
    &
    \textbf{80.9} &
    \textbf{19.2}
    \\
    + Token-based FT &
    60.9 &
    5.2 &
    &
    72.1 &
    14.0
    \\
    \bottomrule
    \end{tabular}
    \caption{Effectiveness of different language control mechanisms on unseen target languages 
    compared to the base model without fine-tuning.
    Instruction-based language control outperforms token-based control.\label{tab:chap6_langcontrol_instruction_vs_token}}
\end{table}

\paragraph{Isolating Language Control Mechanisms}

To test the previous hypothesis that natural language instructions facilitate language control, 
we conduct a controlled experiment by fine-tuning Qwen2.5-0.5B using the same token-based language specification format as M2M-124, 
as described in \S\ref{subsec:data}.
This format eliminates natural language instructions entirely, 
allowing fairer comparisons between models while holding the language control method unchanged.
The results support our hypothesis that instruction-following paradigms provide superior language control. 
As shown in \autoref{tab:chap6_langcontrol_instruction_vs_token}, 
when trained with token-based language control, 
Qwen2.5's performance on unseen target languages drops substantially from 65.3 to 60.9 COMET over 4 unseen target language pairs. 
To account for low initial performance in some non-English language pairs, 
we specifically examine the English-Indonesian pair, which has a stronger baseline.
In this case, performance still degrades substantially from 80.9 to 72.1 COMET and from 19.2 to 14.0 BLEU.
These results on Qwen show that it is the instruction-following ability, 
rather than the decoder-only architecture,
that provides stronger protection against target language forgetting.

\paragraph{Impact of In-Language Instructions}

Building on our previous findings regarding instruction-following for language control, we investigate whether using instructions in the target language (in-language instructions) can mitigate catastrophic forgetting on unseen language pairs. 
While prior work on in-language instructions for multilingual LLMs shows mixed results \citep{marchisio-etal-2024-understanding,mondshine-etal-2025-beyond-english, liu-etal-2025-translation, DBLP:conf/iclr/RomanouFSNSMACH25,enomoto-etal-2025-fair}, these studies primarily evaluate models out-of-the-box. 
In contrast, we focus specifically on the training effects of in-language instructions.

We focus on the Llama3-8B trained on the \textsc{small} dataset, 
which exhibits strong catastrophic forgetting (rightmost plots in \autoref{fig:gain_forget_plot_alma}). 
As the results in \autoref{tab:chap6_inlang_instruct} suggest, 
for unseen language pairs affected by forgetting, in-language instructions substantially outperform English instructions. 
Specifically, average language accuracy improves dramatically from 22.1\% to 82.0\%, 
with corresponding translation quality gains as measured by COMET increasing from 57.2 to 70.9. 
It is worth noting that this does not impact performance on supervised language pairs, and slightly improves performance on unseen target languages (COMET 79.6$\rightarrow$80.3).


\begin{table}[h!]
    \small
    \centering
    \setlength\tabcolsep{1.7pt}
    \begin{tabular}{@{}llcccc@{}}
    \toprule
     & \textbf{Metric} & \textbf{Supervised} & \textbf{\makecell[c]{Unseen\\Pair}} & \textbf{\makecell[c]{Unseen\\Source}} & \textbf{\makecell[c]{Unseen\\Target}} \\
    \midrule
    \multirow{3}{*}{\textbf{Original}} 
    & COMET     & 76.9 & 63.8 & 81.1 & 71.8 \\
    & BLEU      & 26.0 & 10.9 & 29.5 & 14.8 \\
    & LangID    & 97.9 & 85.1 & 97.9 & 93.5 \\
    \midrule
    \multirow{3}{*}{\textbf{\makecell[l]{English\\instruction}}}
    & COMET     & 81.4 & 57.2 & 82.8 & 79.6 \\
    & BLEU      & 30.0 & \phantom{0}4.4 & 31.8 & 15.5 \\
    & LangID    & 96.7 & 22.1 & 98.4 & 94.1 \\
    \midrule
    \multirow{3}{*}{\textbf{\makecell[l]{In-language\\instruction}}}
    & COMET     & 81.7 & 70.9 & 82.9 & 80.3 \\
    & BLEU      & 30.3 & 14.4 & 32.4 & 16.1 \\
    & LangID    & 97.5 & 82.0 & 98.5 & 95.1 \\
    \bottomrule
    \end{tabular}
    \caption{With Llama3 on the \textsc{small} dataset, in-language instructions recover catastrophic forgetting on unseen pairs, reversing a 6.6 COMET loss (63.8$\rightarrow$57.2) into a 7.1 COMET gain (63.8$\rightarrow$70.9).\label{tab:chap6_inlang_instruct}}
\end{table}

\subsection{Analyses by Language Pair Types} \label{subsec:chap6_analyses_lang_pair}
The results in \autoref{fig:gain_forget_plot_alma} and \autoref{fig:gain_forget_plot_wmt21} also suggest that catastrophic forgetting patterns are strongly dependent on the language pair type.
As shown in the previous section, a major issue for language pairs unseen during fine-tuning is generating incorrect output languages. 
Therefore, we separately discuss the two language control mechanisms.

\paragraph{Token-Based Control and Target Language Forgetting}
For translation-specific models (M2M-124 variants) which use specialized tokens for language control, performance degradation is most acute for unseen target languages. 
This is expected, as if the language token for a target language is never encountered during fine-tuning, 
the model's ability to interpret it and generate the correct language catastrophically degrades.

\paragraph{Unseen Pairs as Main Vulnerability for Instruction-Following Models}
In contrast, instruction-following models demonstrate greater resilience on unseen target languages, a capability we attribute to the generalizable nature of natural language prompts (\S\ref{subsec:chap6_architecture_and_lang_control}). 
However, these models are not immune to forgetting and are most susceptible when handling unseen language pairs, where both source and target languages are absent from the fine-tuning set. 
This is particularly evident with the Llama3-8B model.
We hypothesize this vulnerability is compounded by the fact that these unseen pairs are often non-English-centric. 
Base models typically possess weaker zero-shot capabilities for such translation directions 
due to the prevalence of English in their pre-training data. 
Fine-tuning on a different, often English-centric, set of pairs appears to accelerate the forgetting of these already fragile, non-English-centric translation abilities.

\subsection{Comparing LoRA and Full Fine-Tuning} \label{subsec:lora_vs_full_FT}
We observe that LoRA and full fine-tuning result in comparable levels of catastrophic forgetting (fourth and fifth columns of \autoref{fig:gain_forget_plot_alma} and \autoref{fig:gain_forget_plot_wmt21}). 
Note that we applied LoRA adapters to all  components of self-attention and linear projections, 
thereby minimizing differences in parameter coverage as a confounding factor.
Our finding differs from that of \citet{biderman2024lora}, 
who observed that LoRA mitigates forgetting when adapting models to dissimilar domains like code and math. 
We hypothesize that this difference is because our fine-tuning task (translation) requires a smaller domain shift for the base models, 
which already exhibit strong zero-shot translation capabilities, 
whereas adapting to code or math requires a larger deviation.

\section{Evaluating Cross-Lingual Alignment for Forgetting Mitigation} \label{sec:chap6_cross_lingual_align}
Having identified the architectural and training factors that impact catastrophic forgetting, 
we pose a question about mitigation strategies: 
Do established forgetting mitigation methods primarily restore lost performance, 
or do they also improve cross-lingual transfer?
We focus on cross-lingual alignment methods, 
as they encourage similar representations for semantically equivalent content across languages,
which could mitigate forgetting.

\subsection{Evaluated Methods}
We evaluate three prominent cross-lingual alignment techniques that encourage shared representations across languages:
\begin{itemize}[nolistsep,leftmargin=*]
\item \textbf{Adversarial language identification} \cite{ganin2016domain,arivazhagan2019missingingredientzeroshotneural}: 
includes an adversarial language classifier that encourages language-agnostic representations by penalizing the model's ability to predict the source language from hidden states.
\item \textbf{Similarity-only loss} \citep{arivazhagan2019missingingredientzeroshotneural,pham-etal-2019-improving}: pulls together translation pairs without negative examples. 
While a naive implementation would lead to representation collapse, 
joint training with the translation loss mitigates this  by maintaining discriminative power for the primary task \cite{duquenne2023sonarsentencelevelmultimodallanguageagnostic}.
\item \textbf{Contrastive loss} \cite{pan-etal-2021-contrastive}: employs a contrastive objective that pulls together representations of translation pairs while pushing apart representations of unrelated sentence pairs.
\end{itemize}
The losses are applied on encoder for encoder-decoder models, and on the middle layers of decoder-only model \cite{liu-niehues-2025-middle}.

\subsection{Translation-Specific Models} \label{subsec:chap6_langid_translation_specific}
\begin{figure}[t!]
    \centering
    \begin{subfigure}{\linewidth}
    \centering
    \includegraphics[width=\linewidth,clip,trim={0 0 0 0}]{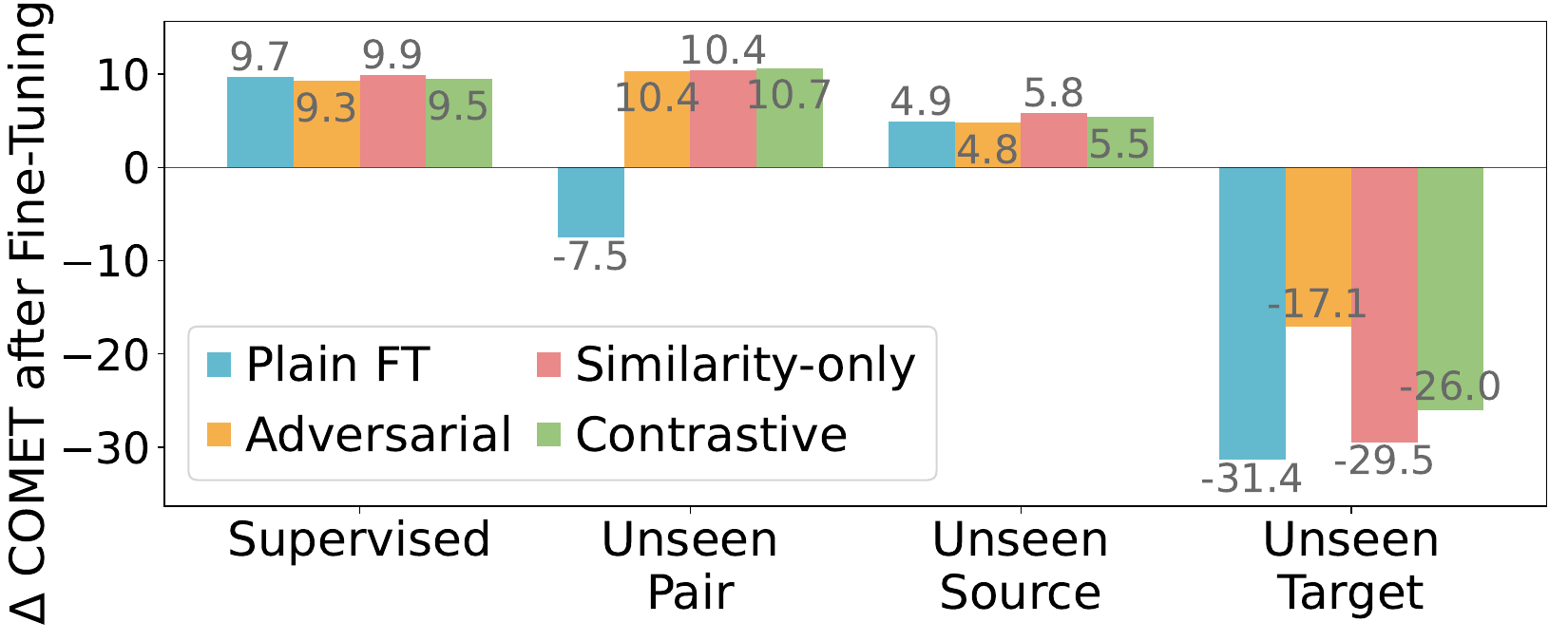}
    \caption{M2M-124-0.2B}
    \label{subfig:barchart_langind_mm100big_wmt21_small_data}
    \end{subfigure}
     ~ 
    \begin{subfigure}{\linewidth}
    \centering
    \includegraphics[width=\textwidth,clip,trim={0 0 0 0}]{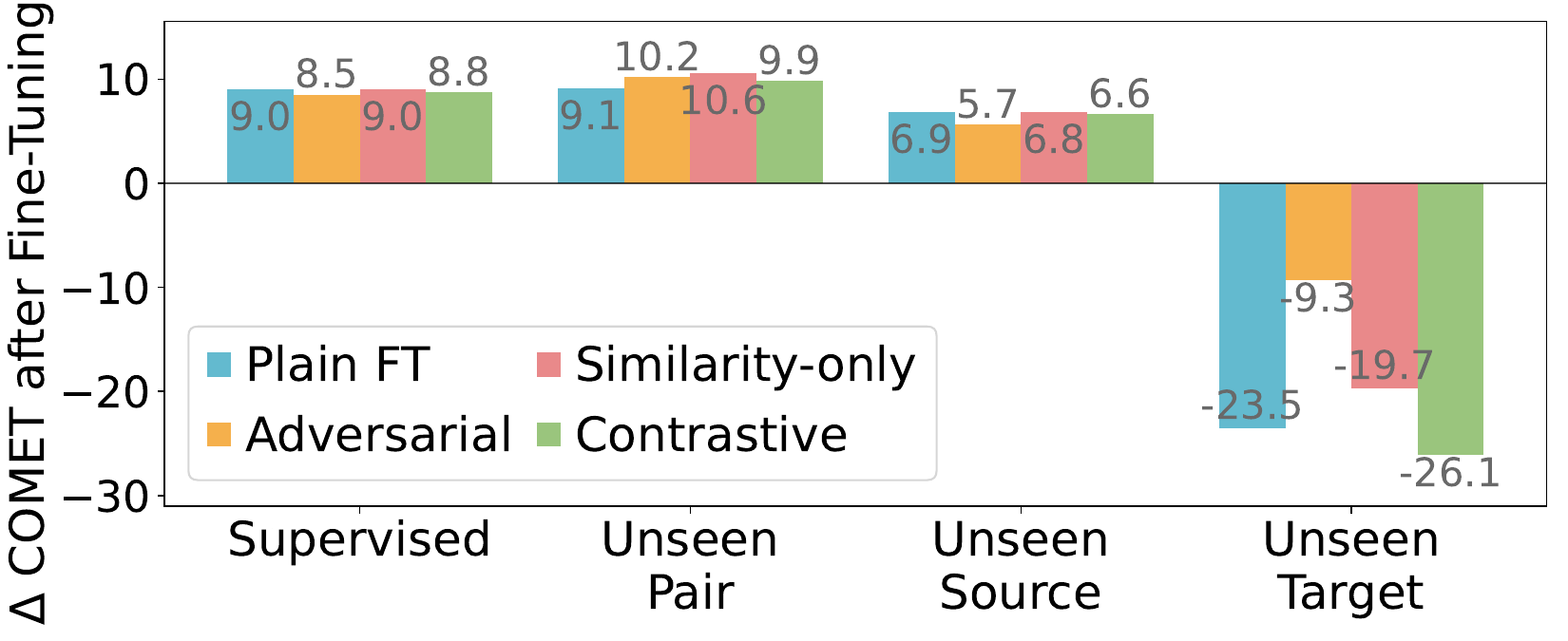}
    \caption{M2M-124-0.6B}
    \label{subfig:barchart_langind_mm100big_wmt21_big_data}
    \end{subfigure}
    \caption{Cross-lingual alignment effects on translation-specific models on the \textsc{large} dataset. Alignment methods bring gains in unseen language pairs, but suffer from persistent forgetting in unseen target languages.\label{fig:barchart_langind_mm100big_wmt21}}
\end{figure}

We first evaluate the three alignment methods on translation-specific models: 
the 0.2B and 0.6B variants of M2M-124.
The results are shown in \autoref{fig:barchart_langind_mm100big_wmt21}, 
displaying change in translation quality for various language categories, 
comparing each alignment method against the plain fine-tuning baseline.

\paragraph{Gains in Unseen Language Pairs} 
Among the three unseen categories, 
alignment methods primarily improve performance on unseen language pairs.
These improvements are observed when plain fine-tuning causes forgetting (\autoref{subfig:barchart_langind_mm100big_wmt21_small_data})
and when it brings improvements (\autoref{subfig:barchart_langind_mm100big_wmt21_big_data}).
For the 0.2B model, 
these methods reverse a -7.5 COMET loss by plain fine-tuning (60.5$\rightarrow$53.0) 
into a gain of over 10 COMET. 
On the larger 0.6B model, the gains are more modest but consistent, ranging from +0.8 to +1.5 COMET over the plain fine-tuning baseline. 
Besides this category, 
alignment techniques do not benefit unseen source or target languages, as discussed next.

\paragraph{Persistent Forgetting in Unseen Target Languages} 
The last column of \autoref{fig:barchart_langind_mm100big_wmt21} 
shows that all three approaches still result in drastic, double-digit COMET degradation for this category. 
This suggests an inherent weakness of the token-based language control mechanism discussed in \S\ref{subsec:chap6_analyses_lang_pair}.
Cross-lingual alignment, while beneficial for transfer, struggle to overcome this fundamental limitation.

\paragraph{Similar Performance Patterns Across Alignment Methods}
The three evaluated alignment methods exhibit highly similar performance patterns. 
While the adversarial approach shows an advantage for unseen target languages (\autoref{fig:barchart_langind_mm100big_wmt21}), 
the improvement is insufficient to overcome the severe forgetting in this category. 
We argue this difference is of limited practical relevance,
as the degradation results from an inherent limitation in the token-based language control that none of the methods fully resolve.
Moreover, the instruction-based language control already demonstrates superior baseline performance in this setting (\S\ref{subsec:chap6_architecture_and_lang_control}).
Therefore, given their comparable overall effectiveness, 
we select a single representative alignment method for the subsequent analysis of instruction-following models.

\subsection{Instruction-Following Models}
\begin{figure}[t!]
    \centering
    \includegraphics[width=\linewidth,clip,trim={0 0 0 0}]{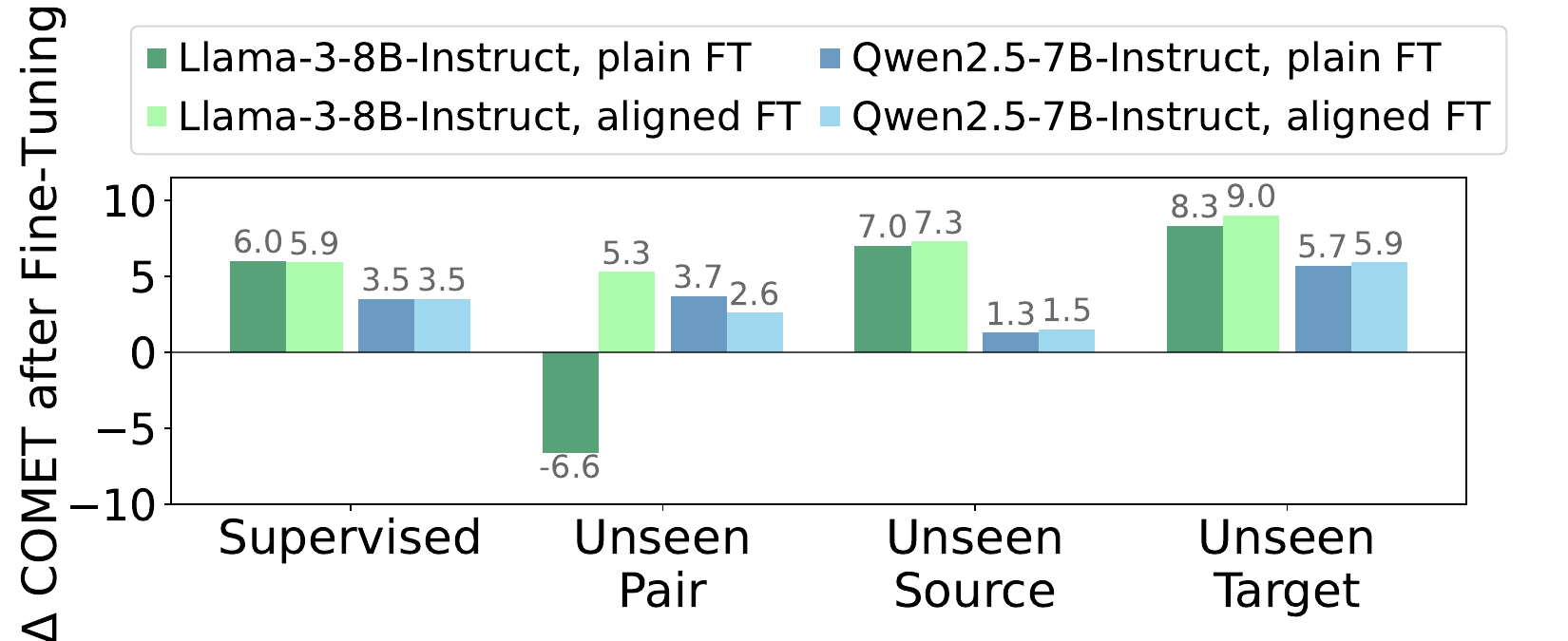}
    \caption{Cross-lingual alignment effects on instruction-following models on the \textsc{small} dataset.\label{fig:barchart_langind_llms}}
\end{figure}

We choose the contrastive approach for studying instruction-following models due to its generality, as the other two approaches require joint training with task-specific loss to avoid collapse. 
In~\autoref{fig:barchart_langind_llms}, results are shown for both Llama3-8B and Qwen2.5-7B with LoRA fine-tuning on both the \textsc{small} and \textsc{large} fine-tuning data configurations.

\paragraph{Impact on Unseen Source and Target Languages}
On unseen source languages, 
cross-lingual alignment generally leads to performance comparable to standard LoRA fine-tuning, in line with previous observations on task-specific models (\S\ref{subsec:chap6_langid_translation_specific}).
On unseen target languages, 
cross-lingual alignment provides a modest gain of 0.7 COMET  (79.6$\rightarrow$80.3) for Llama,
whereas it offers no significant improvement for the Qwen model.
These results suggest that the primary advantage of cross-lingual alignment is its ability to reverse forgetting on unseen language pairs. 
In conditions where standard fine-tuning already yields improvements, the additional gains from alignment are much milder.
This forgetting pattern here, especially in the unseen target category, differs from those observed on translation-specific models in \S\ref{subsec:chap6_langid_translation_specific}. 
The persistent strong forgetting observed previously is substantially reduced, 
with alignment occasionally surpassing the performance of standard fine-tuning. 
This suggests that as models move forward in their instruction-following capabilities, 
their potential for cross-lingual transfer is also enhanced.

\paragraph{Why Gains Concentrate on Unseen Pairs}

The most significant performance improvements are observed on unseen pairs, 
where both the source and target languages were included in the training data but never appearing together.
As discussed in \S\ref{subsec:data}, 
this category is particularly challenging because fine-tuning can cause the model to overfit to spurious source-target associations,
leading to outputs in an incorrect target language.

We interpret these results as evidence that cross-lingual alignment methods directly counteract this degradation. 
Encourage more language-invariant representations leads to disentangling semantic content from language-specific features. 
By breaking the spurious associations learned during training,
alignment mitigates the effects of forgetting and restores the model's ability to generate the correct target language.
Consequently, the performance gains are most substantial on these unseen pairs.
Considering that the number of translation directions in a multilingual system scales quadratically, 
and that many languages may only have parallel data to English,
breaking the spurious correlations that affect unseen pairs is of high practical importance for scalable translation models.

In contrast, for translation directions involving entirely unseen languages, 
the central challenge is a general lack of exposure rather than spurious correlations.
Therefore, the impact of this alignment mechanism is much milder in those scenarios.

\section{Conclusion}
In this work, we aim to resolve ambiguities in the literature regarding when catastrophic forgetting occurs for multilingual fine-tuning for MT. 
Based on our findings, we provide the following practical recommendations: 
\textbf{1)} Consider the relative scale between model size and fine-tuning data. 
Larger datasets may require larger base models to prevent forgetting.
\textbf{2)} Prioritize models with strong instruction-following abilities over specific architectural choices.
\textbf{3)} Do not rely solely on parameter-efficient fine-tuning methods as a forgetting mitigation strategy.
\textbf{4)} For models exhibiting forgetting, cross-lingual alignment is promising for unseen pairs where both source and target languages have been separately seen in fine-tuning.
For instruction-following models,
we recommend training with in-language instructions as an initial data-oriented approach
before proceeding with cross-lingual alignment approaches.

\section*{Limitations}
Our study has several limitations that should be considered when interpreting the results:
\begin{itemize}[nolistsep,leftmargin=*]
\item Our translation experiments focus on English-centric language pairs, which reflects real-world data availability. Extension to non-English pivot scenarios would provide additional validation of our findings' generalizability.
\item While we vary model and data scales systematically, computational constraints limit our exploration to larger size ranges. The dynamics of forgetting in even larger models remain to be investigated.
\item We focus on machine translation as it provides a well-structured testbed for studying multilingual forgetting with clear evaluation metrics. 
Whether similar patterns emerge across other multilingual tasks remains an open question beyond the current scope.
\end{itemize}

\section*{Acknowledgement}
We thank the reviewers for their helpful feedback.
Part of this work was funded by the KiKIT (The Pilot Program for Core-Informatics at the KIT) of the Helmholtz Association. 
The authors gratefully acknowledge the computing time provided on the high-performance computer HoreKa by the National High-Performance Computing Center at KIT (NHR@KIT). This center is jointly supported by the Federal Ministry of Education and Research and the Ministry of Science, Research and the Arts of Baden-Württemberg, as part of the National High-Performance Computing (NHR) joint funding program (https://www.nhr-verein.de/en/our-partners). HoreKa is partly funded by the German Research Foundation (DFG).

\bibliography{custom}

\appendix

\section{Training and Inference Details}
\label{sec:appendix_training_details}
Implementation Frameworks: The M2M-124 experiments were conducted using FairSeq \cite{ott-etal-2019-fairseq}, while Qwen and Llama experiments utilized Hugging Face Transformers \cite{wolf-etal-2020-transformers}.

\paragraph{Training}
For M2M-124, we used a batch size of 16,384 target tokens. For Qwen and Llama models, we used a batch size of 128 sentences.
With M2M-124, we applied a warmup period of 2,500 steps with a learning rate of 1e-4. 
Training was limited to a maximum of 500K updates, with validation runs every 2,000 steps.
Early stopping was triggered if validation loss does not improve for 10 consecutive runs.
With Qwen and Llama, we used a warmup period of 200 steps with a default learning rate of 5e-4. For full fine-tuning of Qwen-7B and Llama-8B, the learning rate was reduced to 1e-4 to due to training instability with higher rates. Validation was conducted every 200 steps, with early stopping applied after 5 consecutive  runs without improvement.
Both model families employed an inverse square root learning rate schedule.

\paragraph{Decoding} During inference, we used beam search with a beam size of 5 for M2M-124 experiments, while greedy search was applied for Qwen and Llama models, following \citet{alves2024tower}.


\end{document}